\documentclass[conference]{IEEEtran}
\IEEEoverridecommandlockouts
\usepackage{cite}
\usepackage{subfigure}
\usepackage{amsmath,amssymb,amsfonts}
\DeclareMathOperator*{\argmin}{argmin}
\usepackage{adjustbox}
\usepackage{graphicx}
\usepackage{textcomp}
\usepackage{enumitem}
\usepackage{xcolor}
\def\BibTeX{{\rm B\kern-.05em{\sc i\kern-.025em b}\kern-.08em
    T\kern-.1667em\lower.7ex\hbox{E}\kern-.125emX}}

\usepackage[utf8]{inputenc}
\usepackage{algorithm}
\usepackage[noend]{algpseudocode}
\usepackage{xparse}
\makeatletter
\NewDocumentCommand{\LeftComment}{s m}{%
  \Statex \IfBooleanF{#1}{\hspace*{\ALG@thistlm}}\(\triangleright\) #2}
\makeatother

\makeatletter
\newcommand{\algmargin}{\the\ALG@thistlm}
\makeatother
\algnewcommand{\parState}[1]{\State%
   \parbox[t]{\dimexpr\linewidth-\algmargin}{\strut #1\strut}}

\usepackage{amsthm}
\theoremstyle{definition}

\newtheorem{theorem}{Theorem}
\newtheorem{corollary}{Corollary}[theorem]

\newtheorem*{remark}{Remark}
\usepackage{multirow}
\algnewcommand\algorithmicforeach{\textbf{for each}}
\algdef{S}[FOR]{ForEach}[1]{\algorithmicforeach\ #1\ \algorithmicdo}
\usepackage{amssymb}

\let\emptyset\varnothing

\begin{document}


\title{Elastic Bulk Synchronous Parallel Model \\for Distributed Deep Learning}

\author{\IEEEauthorblockN{Xing Zhao\IEEEauthorrefmark{1}, Manos Papagelis\IEEEauthorrefmark{1}, Aijun An\IEEEauthorrefmark{1}, Bao Xin Chen\IEEEauthorrefmark{1}, Junfeng Liu\IEEEauthorrefmark{2}, Yonggang Hu\IEEEauthorrefmark{2}}
\IEEEauthorblockA{\IEEEauthorrefmark{1}\textit{Department of Electrical Engineering and Computer Science,}
\textit{York University,}
Toronto, Canada \\
\IEEEauthorblockA{\IEEEauthorrefmark{2}\textit{Platform Computing,}
\textit{IBM Canada,}
Markham, Canada \\
\{xingzhao, papaggel, aan, baoxchen\}@eecs.yorku.ca;\{jfliu, yhu\}@ca.ibm.com}}
}
\maketitle
\begin{abstract}
The \textit{bulk synchronous parallel} (BSP) is a celebrated {\em synchronization model} for general-purpose parallel computing that has successfully been employed for distributed training of machine learning models. A prevalent shortcoming of the BSP is that it requires workers to wait for the straggler at every iteration. 
To ameliorate this shortcoming of classic BSP, we propose \textsc{ElasticBSP} a model that aims to relax its strict synchronization requirement. The proposed model offers more flexibility and adaptability during the training phase, without sacrificing on the accuracy of the trained model. We also propose an efficient method that materializes the model, named \textsc{ZipLine}. 
The algorithm is tunable and can effectively balance the trade-off between quality of convergence and iteration throughput, in order to accommodate different environments or applications. 
A thorough experimental evaluation demonstrates that our proposed \textsc{ElasticBSP} model converges faster and to a higher accuracy than the classic BSP. It also achieves comparable (if not higher) accuracy than the other sensible synchronization models.
\end{abstract}

\section{Introduction} \label{Intro}
The {\em parameter server framework} \cite{dean2012large}\cite{ho2013more} has been widely adopted to distributing the training of large deep neural network (DNN) models \cite{chen2015mxnet}\cite{zhang2017poseidon}. 
The framework consists of multiple \textit{workers} and a logical \textit{server} that maintains globally shared parameters, typically represented as dense or sparse vectors and matrices \cite{li2014scaling}, and it supports two approaches: \textit{model parallelism} and \textit{data parallelism} \cite{chen2014big}. In this paper we focus on data parallelism. 
Data parallelism refers to partitioning (sharding) of large training data into smaller equal size shards and assigning them to workers. Then, the entire DNN model is replicated to each worker. During the training, each worker trains  the  replica  model  using  its  assigned  data shard, sends the locally computed gradients (via \texttt{push} operation) to the server that maintains globally shared parameters (weights) and receives back updated global weights from the server (via \texttt{pull} operation). That {\em weight synchronization step} is critical as it provides to the server a means of controlling the iteration throughput (to boost the {\em convergence speed} in wall-clock time) 
and the quality of convergence (i.e., the {\em accuracy}).

\begin{figure}[!t]
  \centering
  \includegraphics[width=0.483\textwidth]{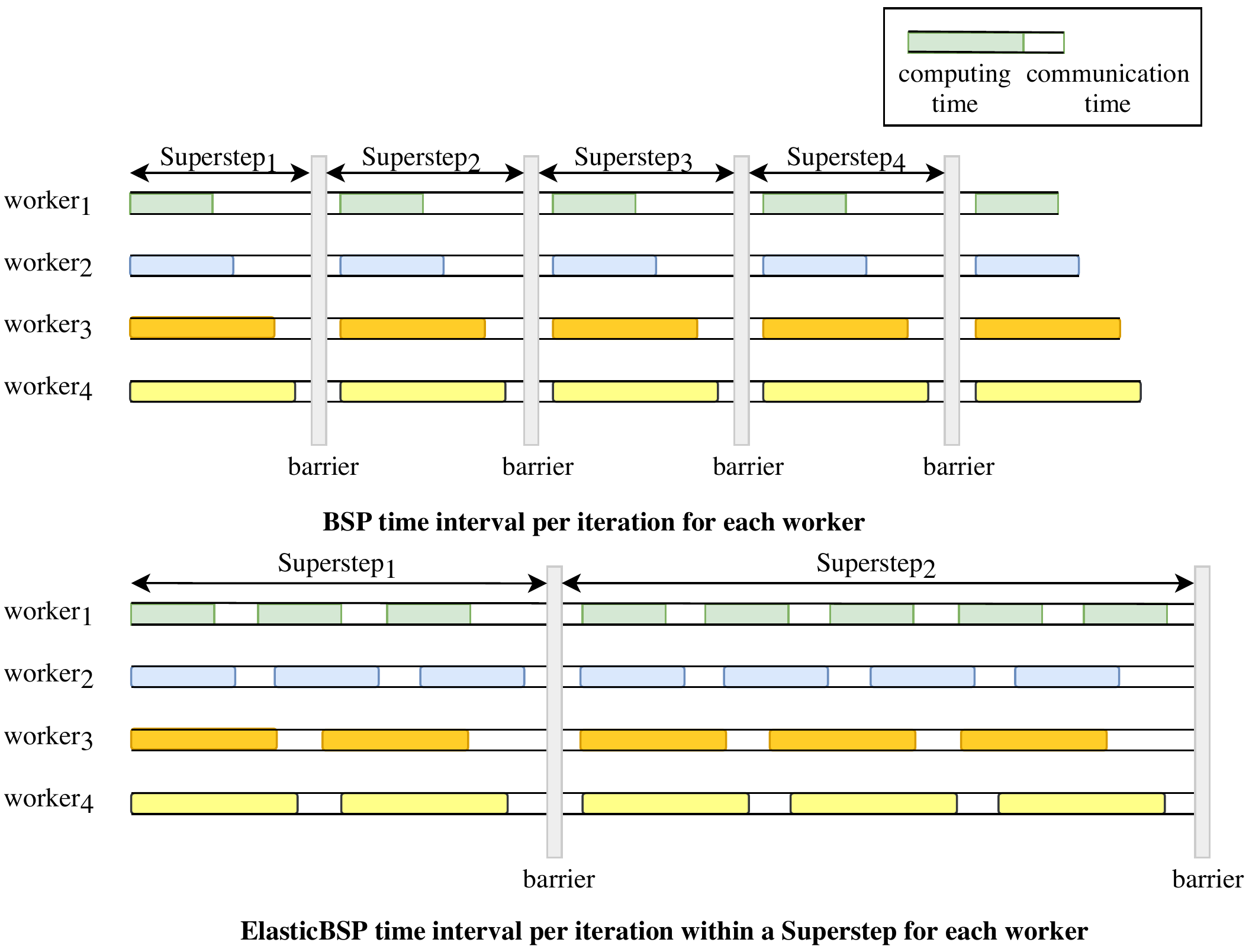}
  \caption{Vanilla BSP and our proposed \textsc{ElasticBSP}. Each {\em barrier} represents the time of weight synchronization among workers and a {\em superstep} represents the time between barriers. In BSP the superstep is fixed to a number of $k$ iterations and all workers have to wait for each other at the end of their $k$ iterations ($k=1$ is shown, which is typical). In \textsc{ElasticBSP}, the time the barrier is imposed varies and each superstep can allow a different number of iterations per worker. These values are determined at runtime by our proposed \textsc{ZipLine} method that achieves minimum overall waiting time of all workers.
  }
  \label{fig:bsp}
\end{figure}

Due to its importance a number of {\em synchronization models} have been proposed, the most important of which are the {\em asynchronous parallel} (ASP), the {\em bulk synchronous parallel} (BSP), and the {\em stale synchronous parallel} (SSP). ASP 
\cite{dean2012large} is the simplest model as it assumes no weight synchronization --- workers always receive different versions of weights from the server at every iteration. BSP \cite{gerbessiotis1994direct} is the most celebrated synchronization model. A critical component of it is the {\em barrier synchronization}, where workers reaching a {\em barrier} have to wait until all other workers have reached it, as well (see Figure \ref{fig:bsp}). During the training phase of a DNN model, each worker, at each iteration, computes the model gradients based on the local data shard and the local weights (originally from the server) and sends the gradients to the server. The server aggregates the gradients of all workers, performs weight update (as one synchronization) and signals the workers to retrieve the latest weights for the next iteration. The workers replace their local weights with the latest weights from the server and start a new iteration. SSP \cite{ho2013more} 
provides an intermediate approach to the two extremes achieved by the ASP and the BSP models. It performs synchronization, but mitigates the strict synchronization requirement of BSP. In principle, it monitors the iteration difference between the fastest and the slowest workers and restricts it to be within a threshold via enforcing synchronization on both workers upon the excess of the threshold.

The aforementioned models exhibit certain limitations. In ASP there is no need for synchronization, so the waiting time overhead of the workers is eliminated. However, the convergence in the training might be dramatically affected due to inconsistent weight updates. On the other hand, a prevalent shortcoming of the BSP is the strict synchronization requirement it imposes. As shown in Figure \ref{fig:bsp}, all workers are waiting for each other by a synchronization barrier. Each {\em barrier} represents the time of the weight synchronization among workers and a {\em superstep} represents the time between subsequent barriers. In BSP-like models the superstep is fixed to a number of $k$ iterations and all workers have to wait for the straggler at the end of their $k$ iterations ($k=1$ is typical), such as in \cite{wangadaptive}. In SSP, while the strict synchronization requirement of BSP is removed, there is still a requirement to manually set the threshold that controls the iteration difference among workers, which remains fixed throughout the training period. Further, SSP does not consider the computational capacity of each worker but merely count on the number of iterations of each worker. 

To ameliorate the shortcoming of current synchronization models, we propose \textsc{ElasticBSP}, a model that aims to relax the strict synchronization requirement of the classic BSP for better convergence. 
Contrary to SSP, the proposed model considers the computational capacity of workers, accordingly, offers more flexibility and adaptability during the training phase, without sacrificing on the accuracy of the trained model. 
The key idea of \textsc{ElasticBSP} is that the time the barrier is imposed varies and each superstep can permit a different number of iterations per worker, offering {\em elasticity} (see Figure \ref{fig:bsp}). We also propose an efficient method that materializes the model, named \textsc{ZipLine}. \textsc{ZipLine} consists of two phases. First, $k$ future iteration intervals (timestamps) of each worker are predicted at run time based on their most recent intervals, assuming a stable environment. Then, a one-pass algorithm operates over the predicted intervals of all workers and performs a lookahead greedy algorithm to determine the next synchronization time (i.e., a time that the overall workers' waiting time overhead is minimized). The algorithm can effectively balance the trade-off between accuracy and convergence speed, in order to accommodate different environments or applications. 
The major contributions of this work are as follows:
\begin{itemize}
    \item we propose \textsc{ElasticBSP}, a novel synchronization model for scaling the training of distributed deep learning models. \textsc{ElasticBSP} replaces the strict synchronization requirement of other BSP-like models with an online decision making about the best time to impose the next synchronization barrier. The model guarantees the convergence for a large number of iterations. 
    \item we design and develop \textsc{ZipLine}, a one-pass algorithm that can efficiently materialize the \textsc{ElasticBSP} model. \textsc{ZipLine} performs online optimization with lookahead to predict the next best synchronization time. 
    It also outperforms sensible baselines. 
    \item we present a thorough experimental evaluation of our \textsc{ElasticBSP} model materialized by the \textsc{ZipLine} on two deep learning models on two popular image classification datasets. The results demonstrate that \textsc{ElasticBSP} converges much faster than BSP and to a higher accuracy than BSP and other state-of-the-art alternatives.
\end{itemize}

The remainder of the paper is organized as follows. Section \ref{background} 
introduces our proposed \textsc{ElasticBSP} model and its properties. Section \ref{problem} formally defines the problem of interest. 
In Section \ref{methodology}, we present algorithmic details of sensible baselines and our proposed method \textsc{ZipLine} to materialize \textsc{ElasticBSP}. Section \ref{experiment} presents an experimental evaluation of the methods. We review the related work in Section \ref{relatedwork} and conclude in Section \ref{conclusion}.

\section{Elastic Bulk Synchronous Parallel Model}
\label{background}
In this section, 
we propose a novel synchronization model that has the premise to ameliorate drawbacks of current models, without sacrificing their benefits.

\begin{figure}[!t]
  \centering
  \includegraphics[width=0.483\textwidth]{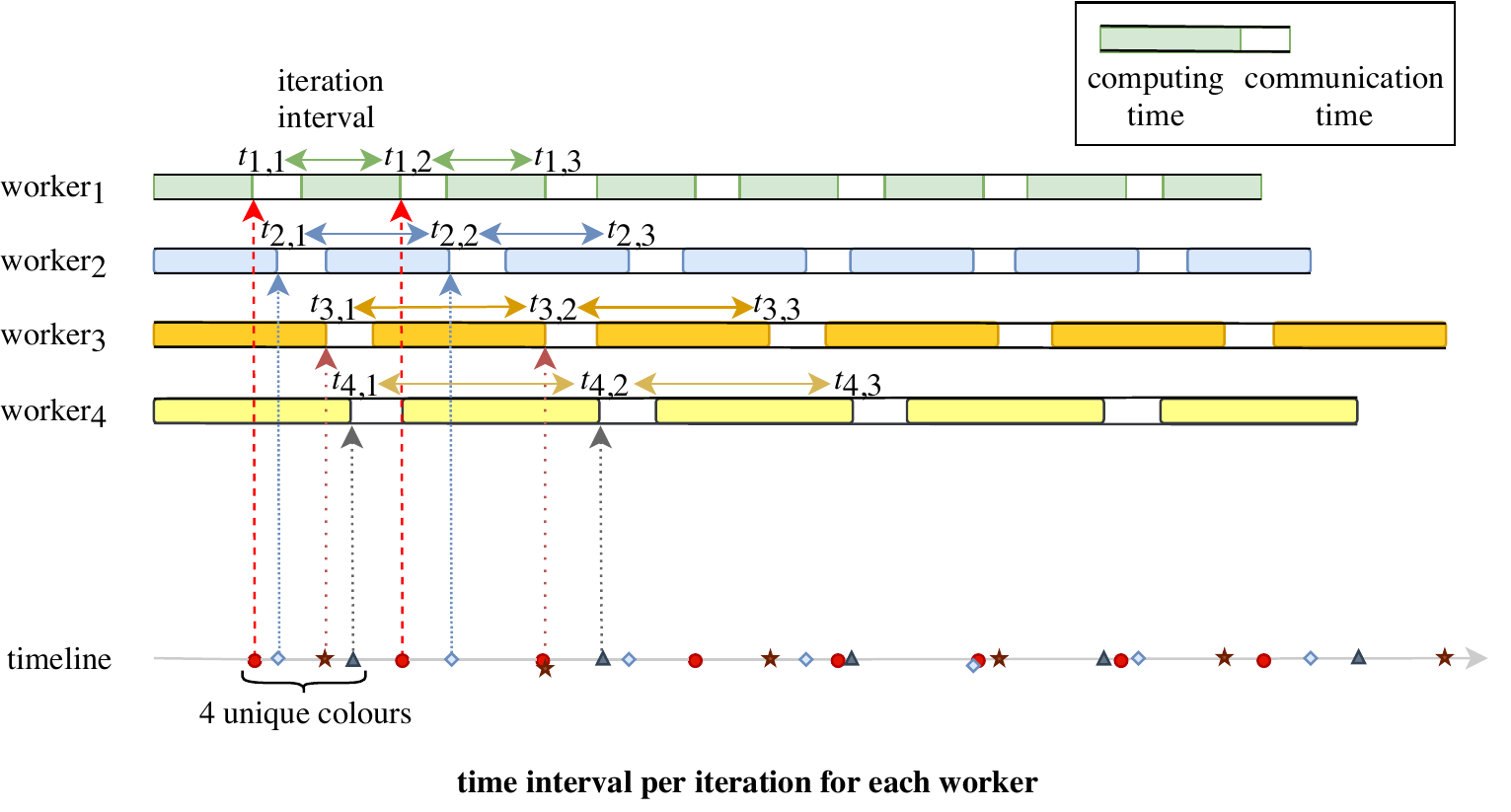}
  \caption{Iteration intervals measured by timestamps of push requests from workers. A dotted line represents the time a push request arrives at the server from a worker. An iteration interval consists of gradient computing period (solid block) and communication period (blank block). All workers' ending timestamps can be mapped onto a timeline. Each timestamp on the timeline is associated to one of the workers. A set which is represented by the bracket always keep $n$ unique values (colors) of workers. \textsc{ZipLine} scans the points from left to right on the timeline, takes one color point into the set per iteration.}
  \label{fig:iterationsSweepLine}
\end{figure}

The BSP model guarantees the convergence on training the DNN models since it is logically functioning as a single server. However, it introduces a large {\em waiting time overhead} due to having to wait for the slowest worker in every single iteration (a mini-batch). On the other hand, the ASP model does not perform any synchronization, so  waiting time for synchronization is minimal, however, it is risky to be used due to its asynchronous scheme that renders the convergence uncertain \cite{zhou2018distributed}. The SSP model offers an intermediate solution to the above two extremes. It guarantees the convergence \cite{ho2013more} when the number of iterations is large and the user specified threshold $\beta$ is small. However, it depends on manually fine-tuning the $\beta$ hyper-parameter which is non-trivial. 

Motivated by the limitations of the current state-of-the-art synchronization models, we propose \textsc{ElasticBSP}. \textsc{ElasticBSP} aims to relax the strict synchronization requirement of BSP. The key properties of \textsc{ElasticBSP} are the following:
\begin{itemize}
    \item The server deals with sequential decision making regarding the {\em best} time that the next synchronization barrier should be imposed (a time when the minimum waiting time for the entire system is achieved). The decision is based on a prediction model that utilizes information about the most recent time interval of each worker available to the server to predict its future intervals. The prediction is based on an online optimization with lookahead and assumes a specific limit $R$ on how many future intervals for each worker should be considered. The need for a specific limit comes from the need to control the algorithm's run time, since that can increase exponentially as the lookahead limit $R$ increases.
    \item The convergence guarantee of the model follows the theoretical analysis of SSP \cite{ho2013more}, where a small iteration difference $\beta$ exists in some period $\tau$ (a superstep). In the case of \textsc{ElasticBSP}, the iteration difference is bounded by the lookahead limit $R$ in some period $\tau$ that is defined by the next best synchronization time. By the end of the period $\tau$, the synchronization barrier is posed to all the workers where gradients aggregation is carried out on the server, similarly to BSP, the weights are synchronized.
\end{itemize}

\textsc{ElasticBSP} offers elasticity in the sense that the distance between two consequent synchronization barriers is not fixed, but it is determined online. In addition, the waiting time is not determined by a fixed iteration difference between the fastest and the slowest workers (as in SSP), but based on the optimal time to synchronize in order to minimize the waiting time. Moreover, the synchronization time is always bounded within the lookahead limit $R$, so it will not simulate the ASP model.

\section{Problem Framework}\label{problem}
Most data centers follow the high availability 
criteria practise \cite{benz2013dependability}, it is realistic to assume that the cluster is running in a stable environment where each iteration time interval (including batch processing and gradient computing) of a worker is similar in a short period. If the worker is not responding in a reasonable time, it will be taken out from the distributed system (and the algorithm in our case). Note that our algorithm is orthogonal to the fault torrent problem. 
Then, we can heuristically predict the future iteration intervals for workers (see Figure \ref{fig:iterationsSweepLine}
) based on their most recent iterations.

\smallskip\noindent\textbf{The Problem}. For $n$ workers in a cluster, each worker $p$ has to process many iterations in a training where each iteration time interval on the same worker $p$ is similar. Each iteration interval is measured by the starting and the ending timestamps of processing an iteration. Suppose we predict $R$ future iterations for each worker. For any worker $p$, it has a set $S^p$ containing a list of starting and ending timestamps of iterations. Most of both timestamps are overlapped for the subsequent iterations. Thus, we only need to use the ending timestamps $e_{i-1}^p, e_i^p$ to measure an iteration $i$. Mathematically we define the set $S^p = \{e_1^p, e_2^p,..., e_R^p\}$ where $e_i^p, i \in [1, R]$ stands for an ending timestamp of worker $p$ and $p \in [1, n]$. The set $S^p$ contains $R$ iterations of worker $p$. We need to find a set $Z$ containing $n$ ending timestamps, one from each set $S^p$, are closest to each other on the timeline. The maximum and minimum difference of these $n$ ending timestamps is the waiting time for a synchronization. 
The smallest timestamp indicates the time-spot for the fastest worker starts waiting whereas the largest timestamp indicates the synchronization barrier to which all workers have to stop for the synchronization. 

From each of these sets $S^p, p \in [1, n]$, we pick one element $e_j^p, j \in [1, R]$ to form a new set $Z = \{e_j^p \}, p\in [1,n]$. The difference between the maximum and the minimum numbers of the set $Z$ is defined as $d_Z = \max(Z) - \min(Z)$. The slowest worker and the fastest worker finish their current iteration at time $\max(Z)$ and $\min(Z)$ respectively. $d_Z$ is the waiting time of the fastest worker. 
Thus, $d_Z$ dominates the overall waiting time for a synchronization since other workers' waiting time were overlapped by the fastest worker's. We are looking for the optimal set $Z^*$ 
which gives the minimum $d_{Z^*}$ from all possible combinations of $Z$. Hence, our objective function is:
\[Z^* = \argmin_Z d_Z\]
\section{Methodology} \label{methodology}
To solve the proposed problem, we first investigate the brute force approach. We analyze the naive brute force searching, \textit{naive search} and develop an optimized version of brute force algorithm named \textit{FullGridScan} since it is infeasible to implement the \textit{naive search} as scaling the number of workers. We next introduce our approach \textsc{ZipLine} to bring down the computation complexity. Lastly, we show the computation and space complexity of the two approaches in Table \ref{tab:alg_complexity}.

\smallskip\noindent\textbf{Naive search}.
\label{scales} In order to find the minimum difference $d_{Z^*}$, a straightforward approach is to use Brute Force. It first checks all possible combinations of selecting a single element from $n$ sets where each set $S$ has $R$ elements. There are $(C_1^R)^n$ combinations. Second, it computes their $d_Z$ values and finds the minimum value $d_{Z^*}$ from all $d_Z$ values. The set $Z^*$ which yields the minimum value $d_{Z^*}$ is the object we are looking for. The computation complexity of this approach is $\mathcal{O}(R^n)$. The space complexity is $\mathcal{O}(R^n)$ to hold the $(C_1^R)^n$ combinations.

\smallskip\noindent\textbf{GridScan}. \label{gridscan} An optimized heuristic brute force algorithm (Algorithm \ref{alg:gridscan}) as a basis component for \textit{FullGridScan}. We consider the predicted $R$ iterations' timestamps for $n$ workers form a $n \times R$ matrix $\mathcal{M}$ where each row of the matrix $\mathcal{M}_p$ represents a worker $p, p\in [1,n]$ and each row $\mathcal{M}_p$ has $R$ predicted iteration points (timestamps) $\mathcal{M}_{p,i}=e^p_i, i\in[1,R]$ for worker $p$. \textit{Designate} any point in $\mathcal{M}$, we can always find a point from other rows with the \textit{shortest distance} to it. Let these closest points from other rows along with the \textit{designated point} in  set $Z$ and we obtain $d_Z$. Accordingly, designate a row of points, we can find $R$ sets of $Z$s associated to every point of the \textit{designated row}. Finally, we can find the set $Z^*$ from $R$ sets of $Z$s with the minimum $d_{Z^*}$. To guarantee we do not miss any early point (on the timeline), we designate the row with the minimum (earliest) timestamp (i.e., $\mathcal{M}_{p,1}$) as the designated row to start the search which costs $\Theta(n)$. The total computation complexity is $\mathcal{O}(R^2n)$. The outer loop over the points on the designated row costs $R$ iterations and the inner loop over each points in $n-1$ rows (workers) constructs one combination $Z$ of distinct $p$ \textit{value} points costs $(n-1)\cdot R$ iterations as there is $R$ points per row. During the search, we only need to keep the set $Z^*$ with the minimum waiting time $d_{Z^*}$ per point in the designated row which requires storage space $\theta(n)$. Along with the storage for $Rn$ points, the space complexity is $\mathcal{O}(Rn)$. 

\begin{algorithm}[!t]
\small
\caption{GridScan - search the set $Z^*$ with minimum $d_{Z^*}$}\label{alg:gridscan}
\begin{algorithmic}[1]
\Procedure{min$_d$Set}{$\mathcal{M}$}
\LeftComment{the $n \times R$ Matrix $\mathcal{M}$ with predicted points} \State $Z^* \gets \emptyset$
\LeftComment*{the set $Z^*$ takes $n$ elements with unique worker id $p$, $p \in [1,n]$}
\State $d_{Z^*} \gets \infty$
\parState{find the row $\mathcal{M}_{p_b}$ with the smallest initial time $\mathcal{M}_{p_b,1}$ from set $\{\mathcal{M}_{p,1}\}$} \label{grid_basis} \LeftComment*{$\mathcal{M}_{p_b,1} = \min( \{\mathcal{M}_{p,1}\}), p \in [1,n]$, $\{\mathcal{M}_{p,1}\}$ the first column of $\mathcal{M}$}
\ForEach{point $e \in$ worker $\mathcal{M}_{p_b}$} 
    \State $Z \gets \emptyset$
    \State add $e$ to $Z$
    \ForEach{worker $\mathcal{M}_p \in \mathcal{M}, \mathcal{M}_p \neq \mathcal{M}_{p_b}$}
        \ForEach{point $\mathcal{M}_{p,i} \in \mathcal{M}_p$}
            \State $\mathcal{M}_{p,min} \gets \argmin_{\mathcal{M}_{p,i}} |\mathcal{M}_{p,i} - e|$
            \LeftComment{the shortest distance point to $e$}
            \State add $\mathcal{M}_{p,min}$ to $Z$
        \EndFor
    \EndFor
    \State $d_Z \gets \max(Z) - \min(Z)$ 
    \If{$d_Z < d_{Z^*}$}
        \State $Z^* \gets Z$; $d_{Z^*} \gets d_Z$
    \EndIf
\EndFor
\State \textbf{return} $Z^*$\Comment{the set with $d_{Z^*}$}
\EndProcedure
\end{algorithmic}
\end{algorithm}

\smallskip\noindent\textbf{FullGridScan}. In \textit{GridScan}, $R$ combinations (of $Z$) are constructed and each of which waiting time $d_Z$ is computed. We expect some critical combinations (containing the smaller waiting time $d_{Z}$) may be missed. In order to cover more useful combinations during the search, \textit{FullGridScan} rotates the designated row of \textit{GridScan} in turn to repeat Algorithm \ref{alg:gridscan} without the line \ref{grid_basis} till all $n$ rows (workers) in $\mathcal{M}$ are covered. It rapidly increases the computation complexity to $\mathcal{O}(R^2n^2)$. \textit{FullGridScan} therefore covers $Rn$ combinations in total versus $R$ combinations explored in \textit{GridScan}. The storage complexity however remains the same as \textit{GridScan}.

\begin{figure}[!t]
  \centering
  \includegraphics[width=0.483\textwidth]{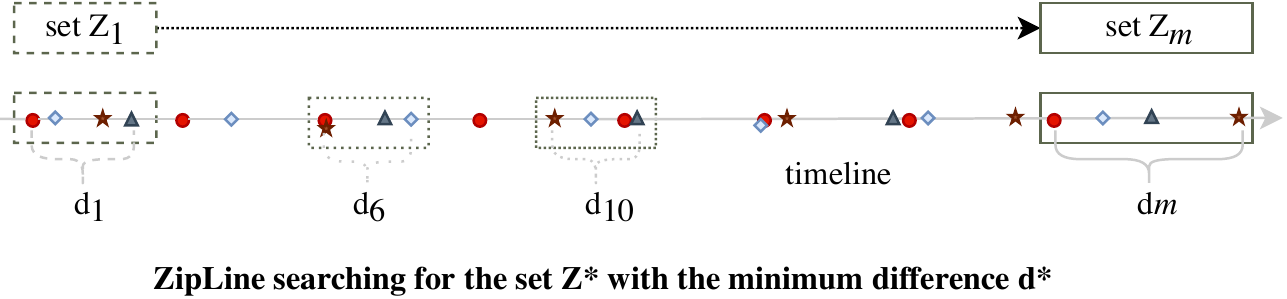}
  \vspace*{-0.5cm}
  \caption{The set $Z$ zips from left to right on the timeline one data point at a time. When $Z$ has $n$ distinct elements, $d_Z$, the difference of minimum and maximum elements of $Z$ is computed. When the set $Z$ reaches to the end of the time line, the minimum $d_Z$ is attained. If multiple minimum $d_Z$s are found, the first minimum $d_Z$ is selected. In the above case, $d_6$ and $d_{10}$ have the same minimum value --- $d_6$ is chosen.}
  \label{fig:zipline}
\vspace*{-0.3cm}
\end{figure}

\noindent{\bf ZipLine.} \textsc{ZipLine} scans through the data points only once in linear complexity $\Theta(Rn)$ as shown in Figure {\ref{fig:zipline}}. In \textsc{ZipLine} (Algorithm {\ref{alg:zipline}}), we first merge all $n$ sets into one large set $\Omega$ and sort its elements in ascending order by their value $e_i^p$ (ending timestamps) where $i \in [1, R]$ and $p \in [1, n]$. We consider the elements are sorted from left to right in position of the set $\Omega$. Second, we define a set $Z$ with the constraint that it contains one timestamp from each worker $p$ at any time as we will use $Z$ to scan every element of $\Omega$ following the timeline from left to right. Intuitively, the set $Z$ only checks the superscript value $p$ of each element $e_i^p$ to prevent duplication of the same worker $p$. If the new timestamp from worker $p$ is added, the old (\textit{duplicate}) timestamp of worker $p$ in $Z$ is removed. Third, we let the set $Z$ scan the set $\Omega$ by iterating one element from $\Omega$ at a time. At the beginning of the scanning procedure, we initialize $Z$ by filling elements from the very left of $\Omega$ to $Z$ while satisfying its constraint till $Z$ has $n$ timestamps from $n$ workers. Then, we compute the minimum and maximum difference (i.e., waiting time) $d_Z$ of $Z$ based on the element value $e_i^p$. Assuming $Z^*$ is $Z$ at the initialization, we store $Z$ to $Z^*$ and $d_Z$ to $d_{Z^*}$. Next, we add one element from the left of $\Omega$ to $Z$ per iteration till $\Omega$ is empty. In each iteration, we compute $d_Z$ and compare its value with $d_{Z^*}$. If $d_Z$ is smaller than $d_{Z^*}$, we store $Z$ to $Z^*$ and $d_Z$ to $d_{Z^*}$. After $Rn$ iterations, we attain the optimal set $Z^*$. The algorithm only uses $\Theta(n)$ space to store $Z^*$. In each iteration, we also iterate through the set $Z$ to remove the \textit{duplicate} element as the new one is added. This operation maintains the invariant (constraint) of $Z$ and costs $\Theta(n)$. Therefore, the total computation complexity is $\mathcal{O}(Rn^2)$ and the space complexity is $\mathcal{O}(Rn)$ for storing $\Omega$. 


\begin{algorithm}[!t]
\small
\caption{ZipLine - search the set $Z^*$ with minimum $d_{Z^*}$}\label{alg:zipline}
\begin{algorithmic}[1]
\Procedure{min$_d$Set}{$\Omega$}\Comment{the merged set $\Omega$} 
\State $Z \gets \emptyset$ 
\LeftComment{the set $Z$ takes $n$ elements with unique $p$ value, $p \in [1,n]$}
\State $\Omega \gets$ sort($\Omega$) \label{sort_bst}
\LeftComment{sort $\Omega$ in ascending order by element's value (timestamp)} 
\While {$|Z| < n$}
\State $\omega \gets$ very left element of $\Omega$ 
\Comment{$\omega$ is $e_i^p$ where $i \in [1, R]$}
\State add $\omega$ to $Z$ 
\LeftComment*{old element of $Z$ is removed if it has the same $p$ value as $\omega$}
\State $\Omega \gets \Omega - \omega$
\EndWhile
\State $d_Z \gets \max(Z) - \min(Z)$ 
\State $Z^* \gets Z$; $d_{Z^*} \gets d_Z$
\While{$\Omega \not= \emptyset$}\Comment{the solution is obtained when $\Omega$ is empty}
\State $Z \gets Z - \min(Z)$ \Comment{$Z$ is in ascending order as of $\Omega$}
\While {$|Z| < n$}
\State $\omega \gets$ very left element of $\Omega$
\State add $\omega$ to $Z$
\State $\Omega \gets \Omega - \omega$
\EndWhile
\State $d_Z \gets \max(Z) - \min(Z)$
\If {$d_Z < d_{Z^*}$}
\State $Z^* \gets Z$; $d_{Z^*} \gets d_Z$
\EndIf
\EndWhile\label{searchMinendwhile}
\State \textbf{return} $Z^*$\Comment{the set with $d_{Z^*}$}
\EndProcedure
\end{algorithmic}
\end{algorithm}
\vspace*{-0.2cm}

\setlength{\tabcolsep}{0.5em}
\begin{table}[!t]
     \caption{Summary of computation and space complexities. 
     }
    \vspace*{-0.1cm}
    \label{tab:alg_complexity}
    \centering
    \begin{tabular}{|l|c|c|}
     \hline
     \bfseries{Algorithm} & \bfseries{Computation} & \bfseries{Space} \\
     \hline
     {\itshape GridScan (heuristic)} & $\mathcal{O}(R^2n)$ & $\mathcal{O}(Rn)$ \\
     {\itshape FullGridScan} & $\mathcal{O}(R^2n^2)$ & $\mathcal{O}(Rn)$ \\
     {\itshape Zipline} &
     {$\mathcal{O}(Rn^2)$} & $\mathcal{O}(Rn)$ \\
     \hline
    \end{tabular}
\end{table}

\section{Experimental Evaluation} {\label{experiment}}
In this section, we run experiments that aim to evaluate: 
\begin{enumerate}[label=\Alph*.]
\item The runtime performance of \textsc{ZipLine} 
to the FullGridScan baseline algorithm. The scalability of \textsc{ZipLine} as a function of the number of workers and the parameter $R$. 
\item The performance of \textsc{ElasticBSP} compared to the classic BSP and other state-of-the-art synchronization models. Which one converges faster and to a higher accuracy? Which one reaches to a fixed number of epochs faster?
\end{enumerate}

\smallskip\noindent\textbf{Dataset}: We generate the datasets 
based on realistic scenarios to evaluate the performance 
of algorithms. 
Table \ref{tab:comparison_ziplines} lists the 
different scales of configurations of datasets for the evaluation. 

\smallskip\noindent\textbf{Environment}: The overhead experiments of \textsc{ZipLine} and baseline algorithms are running on a server with 24x Intel(R) Xeon(R) CPU E5-2620 v3 @ 2.40GHz and 64GB ram.

\subsection{\textsc{ZipLine} Performance Comparison}
In Table \ref{tab:comparison_ziplines}, we 
evaluate the algorithms with 15 predicted iterations for each worker. 
We use 150 predicted iterations to evaluate the scalability of the algorithms to $R$.
The computation time cost of each algorithm is the average of 10 trials.

\setlength{\tabcolsep}{0.5em}
\begin{table}[!t]
    \centering
    \caption{Computation time of algorithms in microseconds/$\mu s$.}
    \vspace*{-0.2cm}
    \begin{tabular}{|l|l|l|l|l|l|l|}
        \hline
         \multirow{2}{*}{Algorithm} & \multicolumn{2}{|c|}{10 Workers} & \multicolumn{2}{|c|}{100 Workers} & \multicolumn{2}{|c|}{1000 Workers} \\
        \cline{2-7}\cline{2-7}
         & R=15 & R=150 & R=15 & R=150 & R=15 & R=150 \\
        \hline
        {\itshape ZipLine} & {\bf 1.49$e$2} & {\bf 1.32$e$3} & {\bf 6.37$e$3} & {\bf 4.99$e$4} & {\bf 2.53$e$5} & {\bf 2.38$e$6} \\
        \hline
        {\itshape FullGridScan} & 1.54$e$3 & 4.67$e$4 & 8.13$e$4 & 2.15$e$6 & 4.04$e$6 & 2.07$e$8 \\
        \hline
        {\itshape GridScan} & 1.68$e$2 & 5.50$e$3 & {\bf\itshape 1.11$e$3} & {\bf\itshape 4.38$e$4} & {\bf\itshape 7.45$e$3} & {\bf\itshape 2.57$e$5} \\
        \hline
    \end{tabular}
    \label{tab:comparison_ziplines}
\end{table}

The combinations of elements from Matrix $\mathcal{M}:n \times R$ increases in exponential as the number of workers $n$ scales or in polynomial as predicted iterations $R$ increments since the combinations is $(C_1^R)^n$ which we described in section \ref{scales}. Table {\ref{tab:comparison_ziplines}} shows that as the number of workers increases the computation time of \textit{FullGridScan} increases much faster than \textsc{ZipLine}. For a fixed number of workers, when the number of predicted iterations per worker increases, the computation time of \textit{FullGridScan} increases much faster than others. 
\textit{GridScan} can be an alternative when the heuristic result is acceptable and the number of workers is larger than 10.

\subsection{Distributed Deep Learning using \textsc{ElasticBSP}}
We compare the performance of \textsc{ElasticBSP} with BSP, SSP and ASP by training DNN models from scratch under each of them on a distributed environment. We set a small threshold $s$=3 for SSP to ensure the convergence and achieve higher accuracy \cite{ho2013more}. For \textsc{ElasticBSP}, we set $R$, the number of predicted future iterations per worker, to 15, 30, 60, 120 and 240 respectively. We ran each experiment three trails and chose the medium result based on the test accuracy.


\smallskip\noindent\textbf{Environment}: 
We implement \textsc{ElasticBSP} 
into MXNet \cite{chen2015mxnet} which supports BSP and ASP models.
The experiments are running on 4 IBM POWER8 machines. Each machine has 4 NVIDIA P100 GPUs, 512 GB ram and 2$\times$10 cores. 

\smallskip\noindent\textbf{Datasets \& DNN models}: We train downsized AlexNet \cite{krizhevsky2012imagenet}, ResNet-50 
\cite{he2016deep} on 
datasets CIFAR-10 and CIFAR-100 \cite{krizhevsky2009learning}. 

\subsubsection{Downsized AlexNet}
We set mini-batch size to 128, epoch to 400, learning rate 0.001 and weight decay 0.0005. \textsc{ElasticBSP} converges faster and to a higher accuracy than other distributed paradigms (see Figure {\ref{fig:downsized_alex}}). BSP converges slower than ASP and SSP but reaches to higher accuracy than both. The increase of $R$ introduces more predicted elements to be computed by ZipLine to determine the optimal synchronization time, therefore, increases the computation overhead. 
As a result, when $R$ becomes larger, it offers nothing but consumes more training time. To this model training, $R$=240 costs extra training time to finish 400 epochs compared to the smaller values. On this model, SSP, ASP, \textsc{ElasticBSP} ($R$=15,30) and BSP complete the fixed 400 epochs in ascending order.

\subsubsection{ResNet-50}
\begin{figure}[!t]
\begin{center}
  \subfigure[\label{fig:downsized_alex}Downsized AlexNet on CIFAR-10 dataset]{\includegraphics[width=\linewidth]{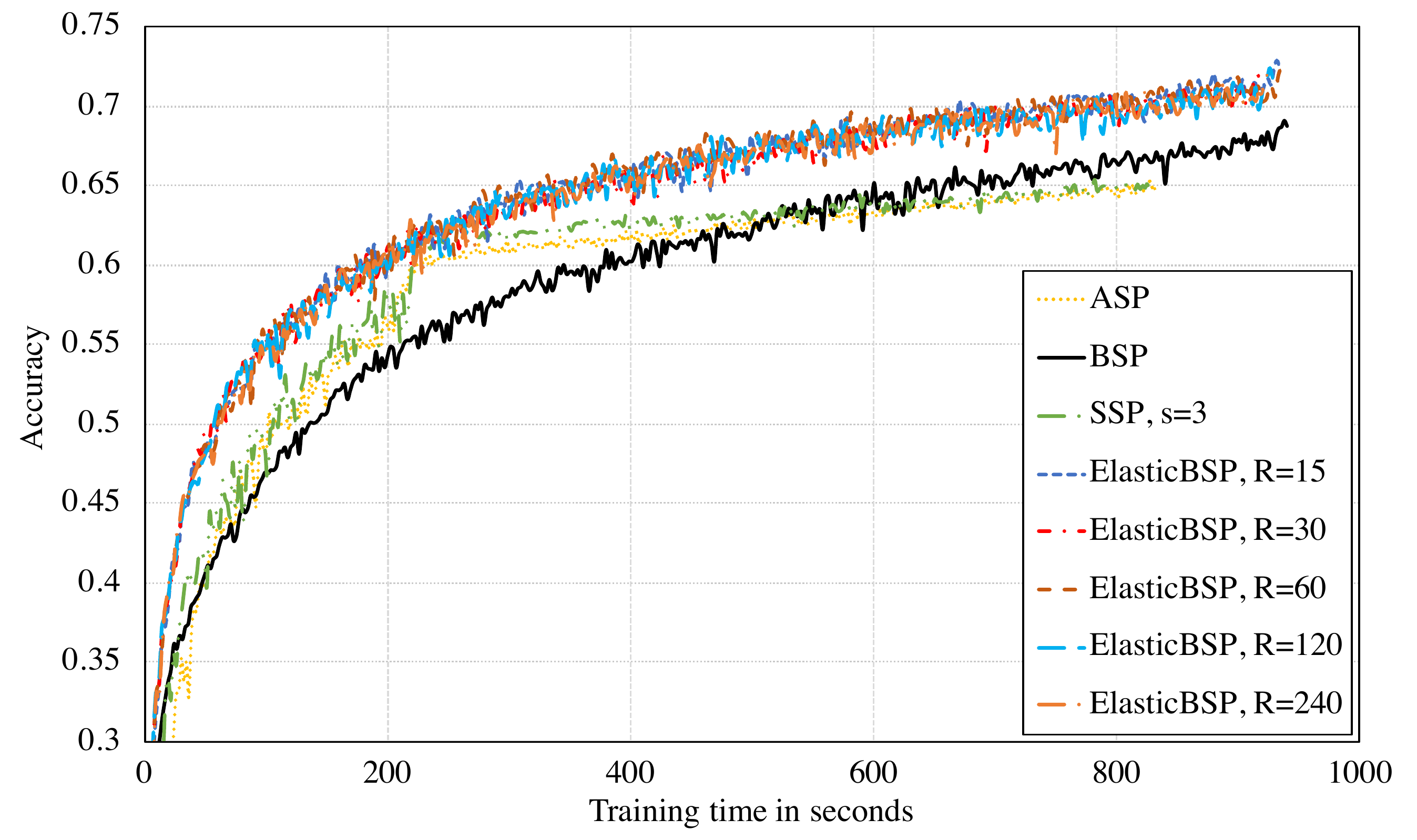}}
  \qquad
  \subfigure[\label{fig:resnet50}ResNet-50 on CIFAR-100 dataset]{\includegraphics[width=\linewidth]{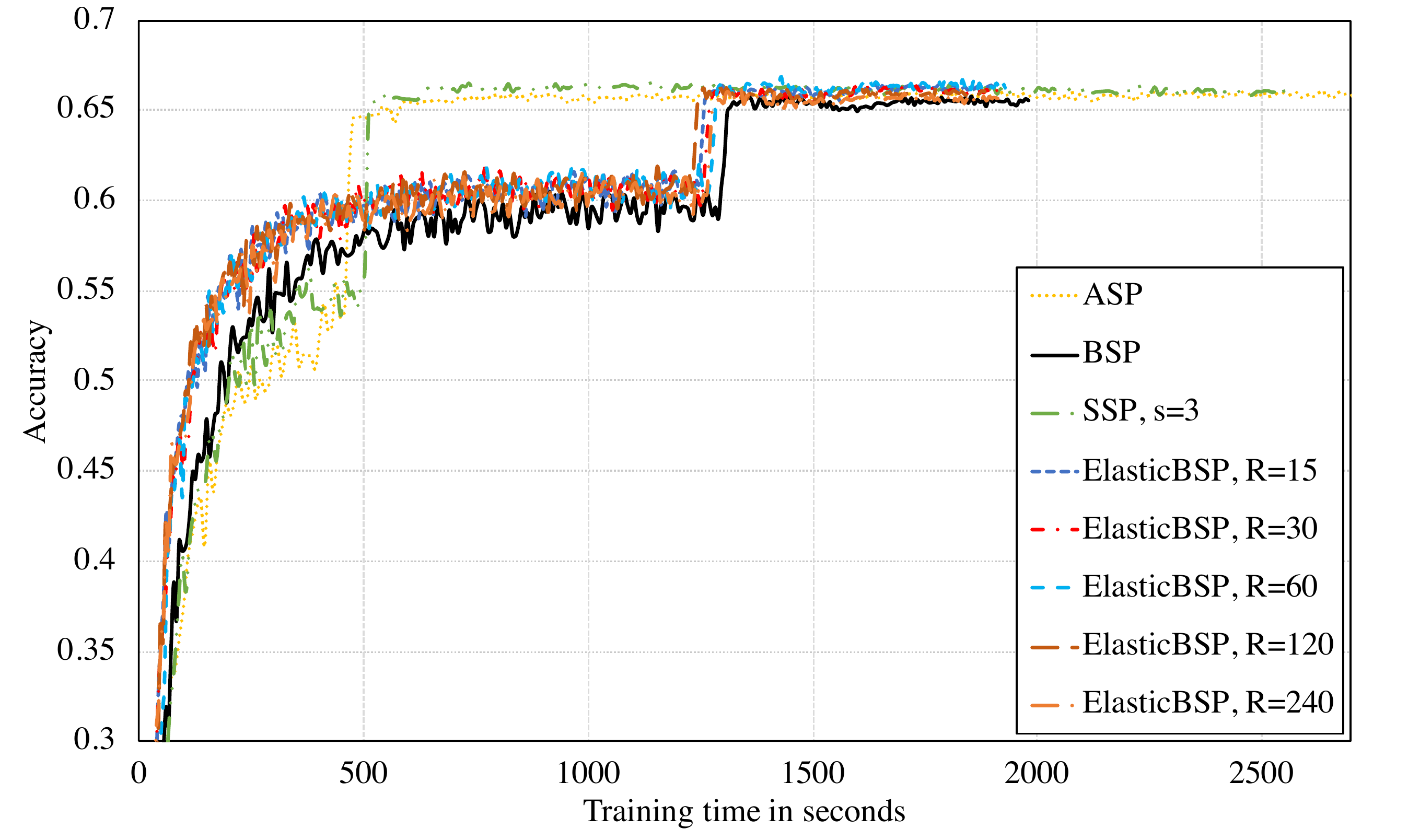}}
\vspace*{-0.5cm}
\end{center}
\caption{Comparison of synchronization models ($n=4$)}
\end{figure}

We set mini-batch size to 128, epoch to 300, learning rate 0.5 and decay 0.1 at epoch 200. 
The results are shown in Figure {\ref{fig:resnet50}
}. 
\textsc{ElasticBSP} converges faster and to a slightly higher accuracy than BSP. Although ASP and SSP converge faster than \textsc{ElasticBSP} and BSP, both cost much more training time to complete 300 epochs. Besides, \textsc{ElasticBSP} converges to a slightly higher accuracy than ASP and SSP. 
ASP and SSP have no bulk synchronization barriers thus have more iteration throughput causing faster convergence. But larger iteration throughput introduces more frequent communications between workers and server and so increases the number of weight updates. However, weight update has to be computed in sequence (as mentioned in Section \ref{Intro}). Thus, their tasks are queued on the server which introduces extra delay. A thorough discussion on why ASP and SSP converge faster but take more training time than BSP can be read in \cite{zhao2019dynamic}. On this model, \textsc{ElasticBSP}, BSP, SSP and ASP complete the fixed 300 epochs in ascending order.

\smallskip\noindent\textbf{Discussion}: 
Above DNN models show that \textsc{ElasticBSP} converges to higher accuracy than BSP and takes less training time when $R$ is not too large. 
Note the different performances of \textsc{ElasticBSP} on the two DNN models are expected since AlexNet contains 2 fully connected layers whereas ResNets has no fully connected layers. Fully connected layers require much less computation time compared to convolutional layers while their representation requires much more parameters than convolutional layers which leads to a large model size. Convolutional networks without fully connected layers such as ResNets takes much more computing time but consumes less communication time due to its smaller model size as to fully connected layer networks. When the ratio of communication time and computation time is small, there is less training time can be saved. More detailed analysis of the different behavior on DNN models with different ratio of computation time and communication time can be read in {\cite{wangadaptive}}. {\cite{zhao2019dynamic}} also provides detailed rationality on the different performances of distributed training using ASP, BSP and SSP on different DNN models.

\section{Related work} \label{relatedwork}
A number of important works closely related to our research has already been cited throughout the manuscript. Here, we elaborate on three alternative models that have been proposed to mitigate the slow down caused by the straggler problem of the classic BSP. A-BSP \cite{wang2018aggressive} handles the straggler problem by terminating the iteration job corresponding to the slowest worker once the fastest workers have completed their jobs. That way, the waiting time is eliminated. The remaining data of the terminated job of the slowest worker is prioritized in the next iteration. This design is limited to the CPU cluster where samples are processed one after another. But in a GPU cluster, a batch of samples are processed all at once in parallel; GPU takes a batch of samples per iteration and computes the gradients. Decreasing the data of a batch (iteration) does not reduce the computation time of GPU. Furthermore, GPU does not support preempt \cite{bauer2011cudadma}. Terminating the job (iteration) means losing all the computed result on that batch of data.
Chen et al. \cite{chen2016revisiting} deal with the straggler problem by adding $k$ extra backup workers to the distributed training with $n$ workers. In this approach, $k+n$ workers are running for the model training. For each iteration, the server only accepts the first $n$ randomly arrived gradient updates from the $n$ faster workers and moves on to the next iteration. The gradients from the $k$ slower workers are dropped. It does save on waiting time of the faster workers but the computing resources of the $k$ slower workers in random iterations are wasted during the training.
ADACOMM \cite{wangadaptive} uses periodic-averaging SGD (PASGD) for bulk synchronization in which workers are doing local updates for $\tau$ iterations before a weight synchronization. That way, the communication time of both uploading gradients and downloading weights from the server per iteration is saved for  $\tau - 1$ iterations. The straggler problem is not addressed in this work. ADACOMM estimates the optimal $\tau$ for a bulk synchronization of local weights based on the training loss. Our \textsc{ElasticBSP} predicts the optimal synchronization time for all workers where each worker has different $\tau$ as opposed to in ADACOMM $\tau$ is uniformly assigned to all workers. 


\section{Conclusion}
\label{conclusion}

In this paper, we proposed \textsc{ElasticBSP} for distributed DNN model training using the parameter server framework. \textsc{ElasticBSP} is relaxing the bulk synchronization requirement of classic BSP and allows asynchronous gradient updates to a certain extent to ensure the quality of convergence and achieve higher accuracy. As a result, it increases the iteration throughput of the workers. \textsc{ElasticBSP} operates in two phases per weight synchronization; first future $R$ iterations for each worker are predicted. Then, \textsc{ZipLine} is applied to determine the optimal next synchronization barrier that minimizes the overall workers' waiting time overhead. \textsc{ZipLine} is a greedy one-pass algorithm 
and adds a minimal overhead on the server, so it can be easily ported in popular distributed machine learning frameworks. The experimental results show that 
\textsc{ElasticBSP} provides faster convergence than classic BSP 
and achieves higher (or comparable) accuracy on the test data sets than other state-of-the-art synchronization models. 

\section*{Acknowledgement}
This work is funded by the Natural Sciences and Engineering Research Council of Canada (NSERC), IBM Canada and the Big Data Research Analytics and Information Network (BRAIN) Alliance established by Ontario Research Fund - Research Excellence Program (ORF-RE).
\bibliographystyle{IEEEtran}
\bibliography{EBSP}
\end{document}